\documentclass{article}
\usepackage{ijcai26}

\usepackage{times}
\usepackage{soul}
\usepackage{url}
\usepackage[hidelinks]{hyperref}
\usepackage[utf8]{inputenc}
\usepackage[small]{caption}
\usepackage{graphicx}
\usepackage{amsmath}
\usepackage{amsthm}
\usepackage{booktabs}
\usepackage{algorithm}
\usepackage{algorithmic}
\usepackage[switch]{lineno}
\usepackage{amssymb}

\usepackage{subfigure}

\theoremstyle{definition} 
\newtheorem{definition}{Definition}

\title{SAGE-LLM: Towards Safe and Generalizable LLM Controller with Fuzzy-CBF Verification and Graph-Structured Knowledge Retrieval for UAV Decision}

\author{
Wenzhe Zhao$^1$\and
Yang Zhao$^1$\and
Ganchao Liu$^1$\and
Zhiyu Jiang$^1$\and
Dandan Ma$^1$\and
Zihao Li$^1$\And
Xuelong Li$^2$\\
\affiliations
$^1$School of Artificial Intelligence, OPtics and ElectroNics (iOPEN), Northwestern Polytechnical University, Xi’an 710072, China\\
$^2$Institute of Artificial Intelligence (TeleAI), China Telecom, China\\
\emails
zhaowenzhe@mail.nwpu.edu.cn,
izhaoyang@nwpu.edu.cn,
xuelong\_li@ieee.org
}

\begin{document}

\maketitle

\begin{abstract}
In UAV dynamic decision, complex and variable hazardous factors pose severe challenges to the generalization capability of algorithms. Despite offering semantic understanding and scene generalization, Large Language Models (LLM) lack domain-specific UAV control knowledge and formal safety assurances, restricting their direct applicability. To bridge this gap, this paper proposes a train-free two-layer decision architecture based on LLMs, integrating high-level safety planning with low-level precise control. The framework introduces three key contributions: 1) A fuzzy Control Barrier Function verification mechanism for semantically-augmented actions, providing provable safety certification for LLM outputs. 2) A star-hierarchical graph-based retrieval-augmented generation system, enabling efficient, elastic, and interpretable scene adaptation. 3) Systematic experimental validation in pursuit-evasion scenarios with unknown obstacles and emergent threats, demonstrating that our SAGE-LLM maintains performance while significantly enhancing safety and generalization without online training. The proposed framework demonstrates strong extensibility, suggesting its potential for generalization to broader embodied intelligence systems and safety-critical control domains.

\end{abstract}

\section{Introduction}

In the field of UAV dynamic decision, traditional methods based on reinforcement learning and optimal control have demonstrated excellent mission completion capabilities in specific scenarios ~\cite{wang2024deep_review}. However, these approaches are typically performance-oriented and prone to hazardous behaviors in complex adversarial settings, with weak safety margins. To address this, safety theories such as reachable analysis~\cite{zhao2025msmar} and Control Barrier Function (CBF)~\cite{sankaranarayanan2024cbf} have been introduced, providing rigorous safety guarantees for control decisions through mathematical constraints. Nevertheless, the safety certification of such methods heavily relies on precise and fixed environmental models. Once encountering unknown obstacles or emergent threats outside the training distribution, their generalization capability declines significantly, making it difficult to balance safety and adaptability in open and dynamic environments~\cite{safaoui2024safe}. Thus, achieving both safety and generalization in open-world UAV pursuit-evasion remains an open challenge.

In recent years, the powerful semantic understanding and zero-shot reasoning capabilities of LLMs have opened new pathways for building generalized autonomous decision-making systems. In robotics, LLMs have been successfully applied to task planning for robotic arms~\cite{mu2023embodiedgpt} and scene understanding for autonomous vehicles~\cite{wang2023drivemlm}, demonstrating their potential in handling open-ended instructions. However, directly applying LLMs to high-dynamic, safety-critical UAV adversarial control still faces two fundamental challenges.

\begin{enumerate}
    \item  Lack of domain-specific knowledge: UAV adversarial control involves in-depth expertise such as aerodynamics and game-theoretic strategies. While general-purpose LLMs possess commonsense reasoning, they lack internalized domain knowledge, and their generated decisions may violate UAV physical constraints or game-theoretic contexts, rendering the decisions semantically plausible yet physically infeasible or even dangerous during execution.
    \item Absence of formal safety guarantees: UAV control imposes extremely high safety requirements, yet the generative nature of LLMs is inherently stochastic and black-box. Their outputs may exhibit inconsistencies and poor interpretability, and more critically, they cannot provide explicit safety certificates, potentially leading to catastrophic failures such as crashes or collisions.
\end{enumerate}

Researchers have conducted a series of studies to address these key challenges. In terms of domain knowledge, approaches mainly include instruction tuning~\cite{zhang2023instruction} and retrieval augmented generation (RAG)~\cite{arslan2024survey}. However, the effectiveness of instruction tuning is limited by the scarcity of high-quality, domain-specific data. Meanwhile, mainstream vector-based RAG methods return isolated text fragments, failing to effectively capture the complex structured relationships among scenario to threat and to strategy of UAV decision, leading to incomplete decision support and weak explainability. Regarding safety verification, existing work predominantly focuses on the linguistic safety of LLM outputs~\cite{ma2025safety}, neglecting the safety of these outputs in embodied control contexts. Some studies incorporate safety modules such as MPC at the low-level control stage~\cite{baumann2025enhancing}, yet these require specific control parameters and suffer from a semantic gap with LLM outputs, making it difficult to establish end-to-end formal safety guarantees.

To systematically address these challenges, this paper proposes the SAGE-LLM towards safety and generalization. The main contributions are summarized as follows:

\begin{itemize}
    \item Star-hierarchical graph RAG for UAV decision knowledge: We introduce a star-hierarchical graph-based RAG to structure UAV decision knowledge across scene, strategy, and control levels with an elastic retrieval method. Compared with traditional RAG method, it provides more coherent, interpretable, and extensible domain knowledge support for LLM-based decision-making.
    \item Fuzzy control barrier function safety verification theory for semantic actions: We define a semantic action cone to model LLM’s discrete instructions as sets with directional uncertainty and accordingly construct a fuzzy CBF. This enables efficient safety certification and automatic correction of high-level semantic actions, establishing for the first time a provable mapping from semantic instructions to safe control inputs.
    \item Experimental validation: We conduct extensive testing in UAV pursuit-evasion scenarios. The results show that SAGE-LLM maintains high task success rates while achieving near-zero collision safety and exhibits strong zero-shot generalization capability. The framework also exhibits strong extensibility and can serve as a reference for broader embodied intelligent systems.

\end{itemize}

\section{Related Work}
\subsection{Safety Research in UAV Adversarial Games}

In UAV pursuit-evasion games, the objective is to accomplish tracking tasks while ensuring safety. Among mainstream approaches, Reinforcement Learning (RL) has demonstrated strong performance in trajectory planning~\cite{zhao2024autonomous}. However, its reward mechanisms often overlook collision risks, leading to insufficient safety in practice. To enhance safety, Safe RL methods incorporate risk penalty terms to suppress unsafe behaviors~\cite{ijcai2024saferl}. Nevertheless, their safety guarantees rely heavily on the distribution of training data, resulting in limited generalization capability. In contrast, formal methods based on safe theories can provide rigorous mathematical safety certificates~\cite{xiao2024safe,he2024agile}. However, their applicability depends on precise system models, making them difficult to adapt to dynamic and open environments.

\subsection{Large Language Models in Embodied Intelligence and UAV Control}

LLMs exhibit strong potential in embodied intelligence due to their semantic understanding and generalization capabilities~\cite{feng2025embodied,liang2025large}. Existing control architectures are primarily categorized into two paradigms: end-to-end and hierarchical. The former suffers from simulation-to-reality gaps and black-box decision-making~\cite{zitkovich2023rt,driess2023palm}, while the latter often exhibits misalignment between semantic instructions and actual control requirements due to a lack of domain-specific knowledge~\cite{ahn2022can,ijcai2025agents}. To improve domain adaptability, research has focused on parameter fine-tuning and knowledge augmentation~\cite{ijcai2025lora,mao2025survey}. However, fine-tuning is constrained by the scarcity of high-quality domain-specific data, and knowledge augmentation through retrieval often yields fragmented results, lacking logical coherence~\cite{wu2025medical}. Furthermore, LLMs still face challenges such as hallucinations and output inconsistency in safety-critical control scenarios. Existing studies predominantly focus on the linguistic safety of LLM outputs~\cite{ijcai2025safe_embodied}, with limited attention to establishing formal guarantees for mapping semantic instructions to safe control actions.

\begin{figure*}[h]
    \centering
    \includegraphics[width=\textwidth]{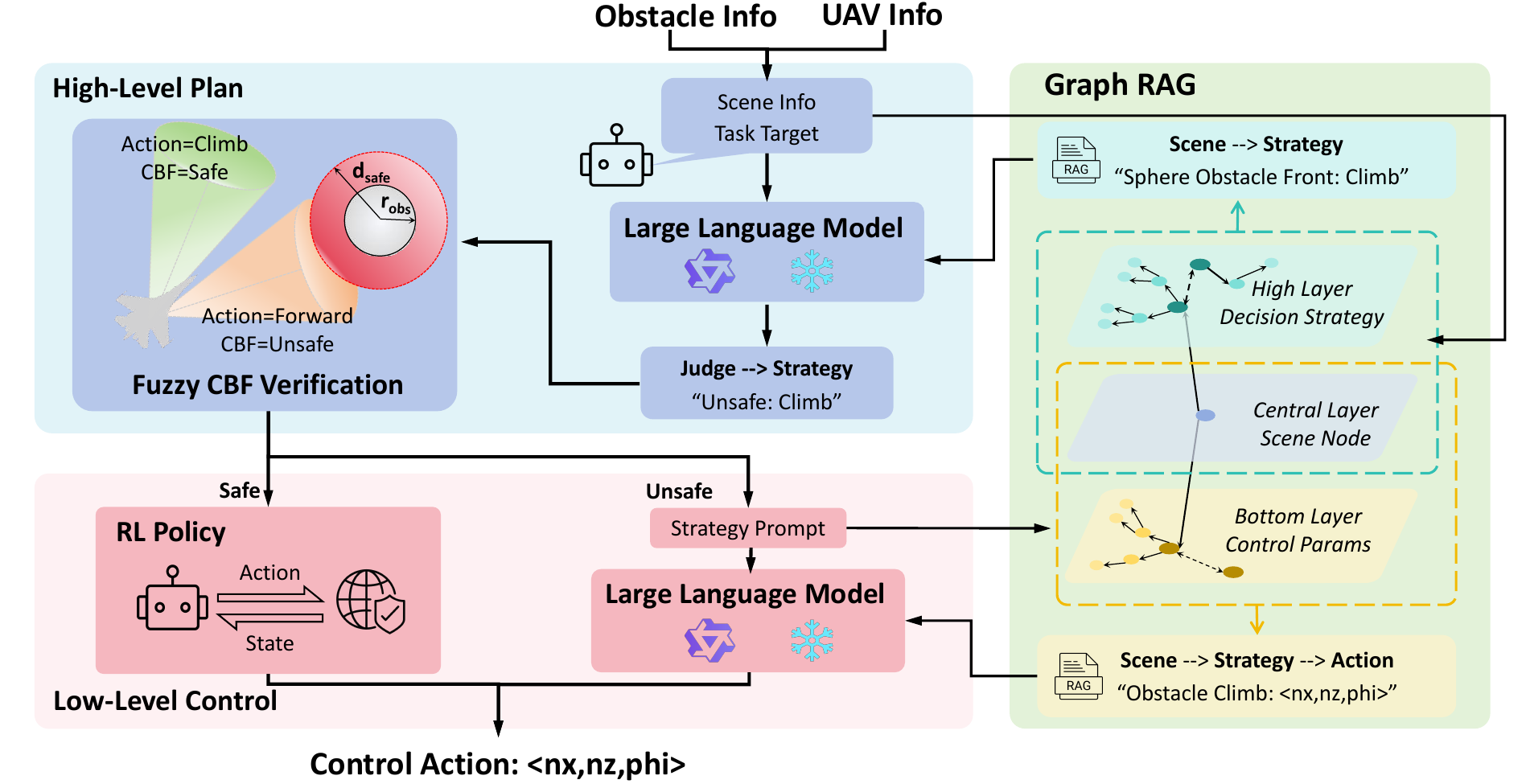}
    \caption{System Framework}
    \label{fig:framework}
\end{figure*}


\section{Preliminaries}
\subsection{Control Barrier Function}
Control barrier functions~\cite{ames2014control} provide a systematic framework for ensuring the safety of dynamical systems by enforcing state constraints. The concept revolves around the forward invariance of a set, ensuring that the system’s state trajectory remains within a predefined safe set over time.

Consider the control-affine system dynamics:
\begin{align}
    \dot{x}=f(x)+g(x)u,\quad x\in\mathbb{R}^n,u\in\mathbb{R}^m,
\end{align}%
where $f(x)$ and $g(x)$ are locally Lipschitz continuous, $x$ represents the system state, and $u$ is the control input. We define the safe set $\mathcal{C}$ as:
\begin{align}
    \mathcal{C} = \{x\in\mathbb{R}^n\mid h(x)\geq0\},
\end{align}%
where $h(x)$ is a continuously differentiable function. The boundary and interior of set $\mathcal{C}$ are given by:
\begin{align}
    \partial\mathcal{C}=\{x\in\mathbb{R}^{n}\mid h(x)=0\},\\ \mathrm{Int}(\mathcal{C})=\{x\in\mathbb{R}^{n}\mid h(x)>0\}.
\end{align}%

The set $\mathcal{C}$ is forward invariant if, for any $ x(0)\in\mathcal{C}$, the solution $ x(t) \in \mathcal{C}$ for all $t \geq 0$. Forward invariance ensures that the system remains within the safe set indefinitely. A function $h(x)$ is called a CBF if there exists an extended class $K_{\infty}$ function $\alpha$ ~\cite{so2024train} such that for all $ x(0)\in\mathcal{C}$:
\begin{align}
    \sup_{u\in\mathbb{R}^m}\left[L_fh(x)+L_gh(x)u\right]\geq-\alpha(h(x)),
\end{align}%
where $L_fh(x)$ and $L_gh(x)u$ are the Lie derivatives of $h(x)$ with respect to $f(x)$ and $g(x)$ respectively which ensures that the set $\mathcal{C}$ is forward invariant under the control law $u$. However, the limitation of conventional CBFs lies in their strict reliance on precise system dynamics models and their requirement that control inputs be continuous low-level signals.

\section{Method}

\subsection{Overall Framework}

SAGE-LLM adopts a two-layer architecture of high-level semantic planning and low-level execution, as illustrated in Figure \ref{fig:framework}. The high-level planner integrates two modules: an LLM responsible for semantic understanding of perceptual information and scene-aware planning, followed by a fuzzy CBF module that performs real-time safety verification of the generated instructions to ensure compliance with formal safety constraints. The low-level controller consists of two components: an RL controller for pursuit-evasion tasks under safe conditions, and an LLM-driven safety controller for precisely obstacle avoidance maneuvers in hazardous states.

Upon receiving perceptual input, the high-level planner conducts scene comprehension, makes high-level strategy and validates safety. If the output is deemed safe, the RL controller is activated to execute the pursuit-evasion task. If the output is classified as hazardous, the LLM safety controller engages, incorporating the high-level avoidance plan to perform safe operations. Throughout this process, an elastic hierarchical graph RAG system provides real-time knowledge retrieval support for both LLM modules, enhancing the the capabilities in domain-specific decision and precise control.

\subsection{Fuzzy CBF Safety Verification Theory}
To ensure the safety of high-level semantic instructions generated by large language models, this section proposes a fuzzy CBF verification theory. First, we establish a geometric directional cone representation model for semantic actions, mapping discrete semantics to continuous control direction sets. Second, a fuzzy CBF verification framework based on this representation is structed, proving the global safety of the directional cone through finite discretization and convex combination.

\subsubsection{Directional Cones for Semantic Actions}
To establish a verifiable mapping from the discrete semantic space to the continuous control space, this paper proposes the directional cone model for semantic actions.

\begin{definition}[Formal Representation of a Semantic Action]
Let $\mathcal{A}=\{a_{1}, a_{2}, \dots, a_{K}\}$ be the finite set of high-level semantic actions. Each action $a_{k} \in \mathcal{A}$ is formally defined as a triple:
\begin{equation}
    a_{k} := (\mathbf{v}_k, \theta_{k}, \mathcal{V}_{k}),
\end{equation}
where:
\begin{itemize}
    \item $\mathbf{v}_k \in \mathbb{R}^{2}$ is the nominal control direction, a unit vector representing the intended motion.
    \item $\theta_{k} \in (0, \pi]$ is the directional uncertainty cone angle, quantifying the maximum allowable deviation during execution.
    \item $\mathcal{V}_{k}$ is the admissible control direction set, defined as the spherical cap:
    \begin{equation}
        \mathcal{V}_{k} := \{\mathbf{v} \in \mathbb{R}^{2} \mid \angle(\mathbf{v}, \mathbf{v}_k) \le \theta_{k}\},
    \end{equation}
    geometrically, $\mathcal{V}_k$ forms a spherical cap region.
\end{itemize}
\end{definition}

This model extends a discrete semantic label into a continuous geometric region, thereby providing the necessary mathematical structure for subsequent formal safety verification based on CBFs. The directional cone angle $\theta_k$ can be parameterized according to the precision of the semantics. For instance, precise tracking may be set with $\theta=10^o$ while emergency evasion may be set with $\theta=60^o$ to allow greater maneuverability.

\subsubsection{Fuzzy CBF Theory Based on Directional Cones}

To establish formal safety guarantees for semantic actions, we first discretize the continuous directional cone $\mathbf{v}_k$ into a set of basis directions. Let $\{ \mathbf{e}_1,\mathbf{e}_2,...,\mathbf{e}_N \}$ be a finite set of basis directions satisfying $\mathbf{v}_k \subseteq conv\{ \mathbf{e}_1,\mathbf{e}_2,...,\mathbf{e}_N \}$, where $conv$ denotes the convex hull, i.e., the set of all convex combinations of the basis directions.~\cite{wang2024surveyconvex}

For each basis direction $e_i$ we define a continuously differentiable safety function $h_i(\mathbf{x})$ and a feasible control law $\mathbf{u}_i$ that satisfy the CBF condition:
\begin{align}
    L_fh_i(\mathbf{x})+L_gh_i(\mathbf{x})\mathbf{u}_i\geq-\alpha_i\left(h_i(\mathbf{x})\right),
\end{align}%
where $\alpha_i$ is a class $\mathcal{K}_\infty$ function.

For any direction $\mathbf{v}\in\mathcal{V}_k$, there exist non-negative coefficients $\lambda_1,\ldots,\lambda_N$ with $\sum_{i=1}^N\lambda_i=1$ such that $\mathbf{v}=\sum_i\lambda_i\mathbf{e}_i$. We then construct the mixed control law and mixed safety function for direction $\mathbf{v}$ as:
\begin{align}
    \mathbf{u_v}=\sum_{i=1}^N\lambda_i\mathbf{u}_i,\quad h_\mathbf{v}(\mathbf{x})=\sum_{i=1}^N\lambda_ih_i(\mathbf{x}).
\end{align}%

Under the above construction, for any $\mathbf{v} \in\mathcal{V}_k$ the mixed control law $\mathbf{u_v}$ satisfies:
\begin{align}
    L_fh_\mathbf{v}(\mathbf{x})+L_gh_\mathbf{v}(\mathbf{x})\mathbf{u_v}\geq-\alpha\left(h_\mathbf{v}(\mathbf{x})\right),
\end{align}
where $\alpha(\cdot)=\max_i\alpha_i(\cdot)$. The detailed provement could be found in Appendix A.

Given $M$ spherical obstacles parameterized by center $o_j$ and radius $r_j$, we introduce the fuzzy CBF criterion for action $a_k$ aggregates the risk over the entire uncertainty cone $\mathcal{V}_k$:

\begin{equation}
\begin{split}
    h_{fuzzy}(\mathbf{x}) = & \min_{\mathbf{v} \in \mathcal{V}_k} \left[ \min_{j \in \{1,\dots,M\}} (\|p(\mathbf{x}, \mathbf{v}) - o_j\| - r_j) \right] \\& - d_{safe},
\end{split}
\end{equation}
where $p(\mathbf{x}, \mathbf{v})$ is the predicted position along the direction of $\mathbf{v}$, and $d_{safe}$ denotes the safety margin. This criterion effectively guarantees safety by bounding the distance to the nearest obstacle under the worst-case directional deviation within $\mathcal{V}_k$.

\subsection{Star-Hierarchical Graph Based RAG}

Conventional RAG paradigms typically treat knowledge as flat embedding spaces, thereby overlooking the inherent hierarchical dependencies, specifically from situational awareness to strategic reasoning and to control execution pipeline within dynamic game-theoretic tasks for UAVs. To address this limitation, this chapter proposes a Star-Hierarchical Graph (SHG) framework for knowledge retrieval. The framework aims to establish a structured, relational, and elastically fault-tolerant mechanism for knowledge representation and invocation.

\subsubsection{Construction of the Star-Hierarchical Graph }
The framework models domain knowledge as a directed star-hierarchical graph $G=(\mathcal{N}, E)$. To balance global decision consistency with local execution flexibility, the node set $\mathcal{N}$ and edge set $E$ are partitioned into three disjoint subsets:
\begin{align}
    \mathcal{N} = \mathcal{N}_S \cup \mathcal{N}_H \cup \mathcal{N}_L, \quad E = E_S \cup E_H \cup E_L.
\end{align}%

The node set $\mathcal{N}$ defines the basic graph structure.
\begin{itemize}
    \item Scene Hub Node ($\mathcal{N}_S = \{n_s\}$): The global context center serving as the logical origin for retrieval.
    \item Hierarchical Subgraphs ($\mathcal{N}_H, \mathcal{N}_L$): $\mathcal{N}_H$ captures semantic evolution from perception to strategy; $\mathcal{N}_L$ maps action categories to execution parameters.
\end{itemize}

The edge set $E$ defines the multi-dimensional paths for knowledge flow:
\begin{itemize}
    \item Connections ($E_S$): This set directly connects the central scene hub $n_s$ to the initial layer nodes of the policy graph $\mathcal{N}_H$ and control graph $\mathcal{N}_L$. It establishes a synchronous constraint from the global context over all subsequent decision branches, 
    \item Intra-layer Logical Associations ($E_H, E_L$): Within each subgraph, edges are further bifurcated into vertical sequential edges $E_{seq}$ and horizontal relational edges $E_{rel}$. $E_{seq}$ connect nodes in adjacent layers to model the causal progression of the decision chain. $E_{rel}$ link semantically similar or logically equivalent nodes within the same layer. 
\end{itemize}

\begin{figure*}[h]
    \centering
    \includegraphics[width=\textwidth]{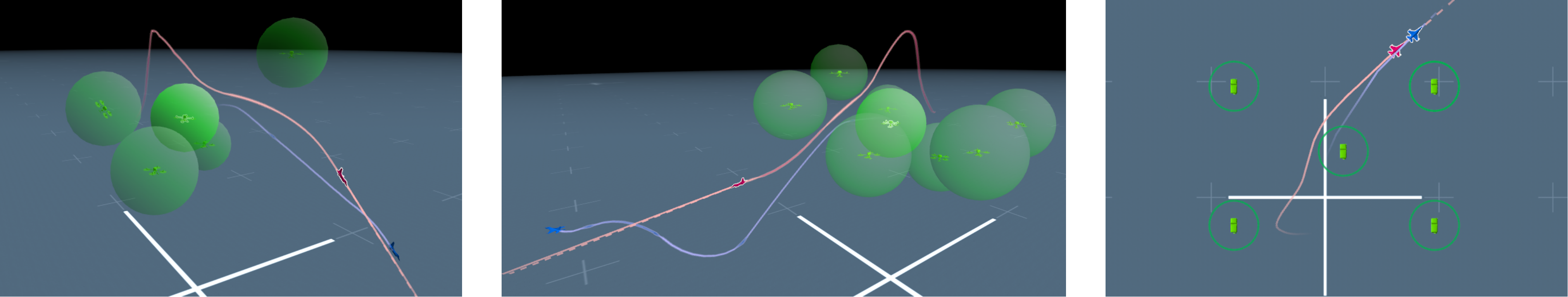}
    \caption{Generalization Experience Visualization. From left to right, the scenarios are: 5 spherical obstacles at random positions, 8 spherical obstacles at random positions, and 5 cylindrical obstacles with infinite height at fixed positions. The results show that SAGE-LLM successfully pursues the target while avoiding all obstacles in unseen, challenging environments, showcasing its zero-shot generalization capability.}
    \label{fig:generalization_result}
\end{figure*}

\subsubsection{Hierarchical Recursive Elastic Retrieval }
To mitigate the semantic ambiguity inherent in UAV game-theoretic queries $Q$, this chapter introduces the Hierarchical Recursive Elastic Retrieval algorithm to make retrieval efficiently. Adopting a neural-symbolic fusion strategy, we define a hybrid matching scoring function $\Phi(n, q_l)$ for node $v$ and query component $q_l$:
\begin{align}
    \Phi(n, q_l) = \alpha \cdot \text{cos}(\mathbf{z}_v, \mathbf{z}_{q_l}) + (1-\alpha) \cdot \mathbb{I}_{syn}(n, q_l),
\end{align}
where $\mathbf{z}$ denotes the semantic embedding vectors extracted via pre-trained language models, and $\mathbb{I}_{syn}$ is the fuzzy matching indicator function. Specifically, $\mathbb{I}_{syn}=1$ if the query value $q_l$ exactly matches the node value or exists within the synonym set $\text{synset}(n)$. By tuning the weight $\alpha$, the algorithm balances the semantic generalization of vector spaces with the rigorous constraints of symbolic domain knowledge.

To enable structured retrieval, the query is defined as a hierarchical sequence $Q = \{q_1, \dots, q_L\}$, where each $q_l$ targets nodes of different semantic levels within the graph layers. During execution, SAGE-LLM performs a top-down, layer-by-layer search to identify an optimal decision path $P = \{n^{(1)}, \dots, n^{(L)}\}$ that maximizes:
\begin{equation}
\begin{split}
    \max_{P} \mathcal{J}(P|Q) = & \sum_{l=1}^{L} \log(\Phi(n^{(l)}, q_l) + \epsilon) \\
    \text{s.t.} \quad & (n^{(l)}, n^{(l+1)}) \in E_{seq} \cup E_{rel},
\end{split}
\end{equation}
where $\epsilon$ is a small positive constant added to ensure numerical stability during the log-likelihood calculation. More details could be found in Appendix B. This hierarchical retrieval mechanism provides precise information via the efficient, elastic search structure that grounds high-level intent into domain-specific, executable control knowledge.

\section{Experience}

\subsection{Experimental Overview}
 The experiments are centered around the task of UAV dynamic pursuit-evasion games, with the core objective of addressing the following four key research questions to validate the framework's effectiveness, advantages, and robustness:
 \begin{enumerate}
     \item Safety Verification: During intense dynamic confrontation, can the SAGE-LLM framework strictly adhere to formal safety constraints and avoid collisions while pursuing mission objectives?
     \item Generalization Capability Verification: Compared to existing mainstream methods, does SAGE-LLM demonstrate superior zero-shot adaptation and decision-making capabilities when dealing with unseen obstacles, sudden threats, and complex environmental layouts?
     \item Architectural Robustness Verification:  Is the core performance of the SAGE-LLM framework independent of the architecture and scale of the base LLM backbone?
     \item Module Contribution Analysis: Through ablation studies, what are the quantitative contributions and necessity of the two core innovative modules ?
 \end{enumerate}

\subsubsection{Experimental Platform}
All experiments are run on NVIDIA RTX 4090 GPU and conducted within the NVIDIA Isaac Gym environment. 

\subsubsection{Evaluation Metrics}
To evaluate the safe and generalized ability, we introduce three matrix referring to ~\cite{zhao2025msmar}.
\begin{enumerate}
    \item Success Rate: The task success rate measures the percentage of missions in which the UAV successfully locks onto the adversarial UAV and completes the task.
    \begin{equation}
        \begin{aligned}
        Rate_{success}=\frac{Num_{success}}{Num_{total}}.
        \end{aligned}
    \end{equation}
    
    \item Safety Rate: The safety rate measures the average probability of violating safety rules (colliding with obstacles) while completing the mission.
    \begin{equation}
        \begin{aligned}
        Rate_{safe}=1-\frac{NumStep_{danger}}{NumStep_{total}}.
        \end{aligned}
    \end{equation}
    Besides, to evaluate the zero-constraint-violation ability of our algorithm, we introduce a zero-danger-rate:
    \begin{equation}
        \begin{aligned}
        Rate_{zero\ danger}=\frac{NumEpisode_{safe}}{NumEpisode_{total}},
        \end{aligned}
    \end{equation}
    where the episode means single whole game.
    
\end{enumerate}

\begin{figure*}[h]
    \centering
    \includegraphics[width=\textwidth]{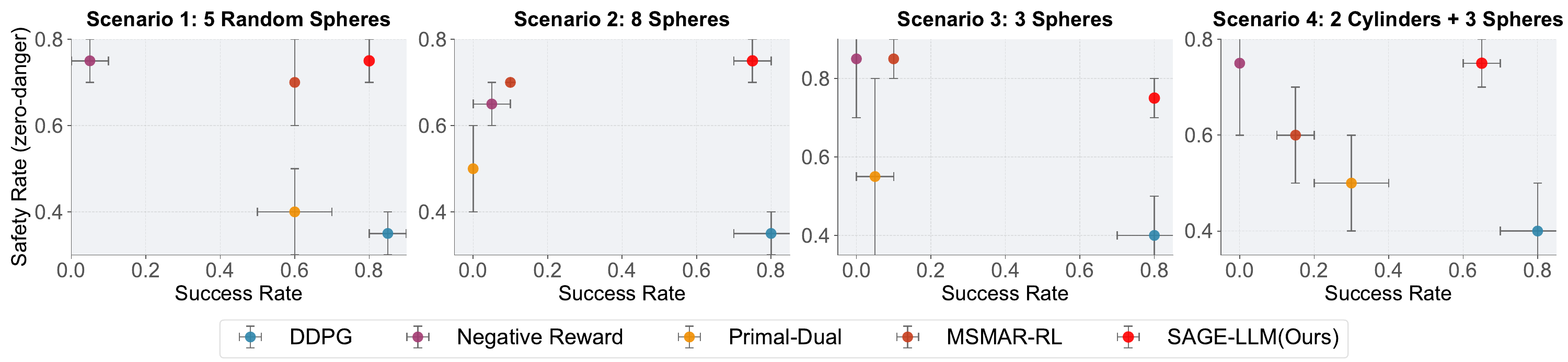}
    \caption{Generalization performance scatter plot with error bars. Proximity to the \textbf{top-right} signifies superior overall performance, while \textbf{smaller error bars} reflect higher stability. SAGE-LLM demonstrates a leading and robust position.} 
    \label{fig:generalization_experience}
\end{figure*}

\subsection{Experiment 1: Safety Verification}
\subsubsection{Experimental Setup}
This experiment aims to validate the core capability of the SAGE-LLM framework in ensuring safety within dynamic adversarial environments. The operational airspace contains five static spherical obstacles with fixed positions and radii, randomly distributed. The pursuing UAV must intercept a target drone moving along a fixed straight line at a constant velocity within this environment and precisely avoid all spherical obstacles while planning its pursuit trajectory.
\subsubsection{Baselines}
For a comprehensive safety assessment, we selected four representative Safe RL methods based on deep reinforcement learning as baselines, and compared them with our proposed SAGE-LLM:

\begin{enumerate}
    \item DDPG: A standard deep deterministic policy gradient algorithm without safety constraints. 
    \item Negative Reward: Incorporating a collision penalty term into the reward function.~\cite{qu2023pursuit}
    \item Primal-Dual Method: RL with constraints using Lagrangain optimization. ~\cite{yue2023research}
    \item MSMAR-RL: A state-of-the-art (SOTA) safe RL algorithm with recovery strategy.~\cite{zhao2025msmar}
\end{enumerate}

\begin{table}[h]
    \centering
    \resizebox{\linewidth}{!}{
    \begin{tabular}{cccc}
        \toprule
        Method          & Success Rate      & Safe Rate     & Zero-Danger-Epoch \\
        \midrule
        DDPG            & \textbf{92\%}     & 90.57\%                   & 52\%      \\
        Punishment      & 56\%              & 92.93\%                   & 60\%      \\
        Primal-Dual     & 60\%              & 97.52\%                   & 48\%      \\
        MSMAR-RL        & 64\%              & 98.29\%                   & 80\%      \\
        SAGE-LLM        & 90\%              & \textbf{99.73\%}          & \textbf{92\%}      \\
        \bottomrule
    \end{tabular}}
    \caption{Baselines Comparison}
    \label{tab:baseline}
\end{table}

\subsubsection{Results and Analysis}
As shown in Table 1, a clear trade-off between task success rate and safety performance is observed across the baseline methods: While the game-theoretic DDPG approach achieves the highest task success rate, it exhibits the weakest safety performance. Other safe reinforcement learning methods, despite improving safety to varying degrees, suffer from a noticeable decline in task success rates. In contrast, the proposed SAGE-LLM framework attains the highest safety rate while maintaining a high task success rate, with a zero-danger epoch ratio of 92\%. These results demonstrate that SAGE-LLM can provide reliable safety guarantees for dynamic adversarial scenarios with almost no loss in task performance.

\subsection{Experiment 2: Generalization Capability}
\subsubsection{Experimental Setup}

Base Scenario: This scenario is used for training the baseline reinforcement learning policies and for constructing the knowledge base. The scene contains five spherical obstacles, whose initial positions are fixed in a standard layout. The target drone employs a matrix game-based evasion strategy.

Generalization Test Suites: All methods are tested directly in the following three categories of generalization scenarios after training/construction on the base scenario, without any fine-tuning:

a) Randomization of Obstacle Positions: The initial positions of all obstacles are randomized to evaluate the system's robustness and zero-shot adaptability.

b) Generalization of Obstacle Quantity: The number of obstacles is varied from the baseline of 5 to either 3 or 8, testing the framework's stability.

c) Generalization of Obstacle Types: Unseen cylindrical obstacles are introduced and mixed with original spheres to simulate real-world heterogeneous obstacles.

\subsubsection{Results and Analysis}

The experimental trajectory visualizations are presented in Figure 2, in which our SEGE-LLM could pusuit the target UAV without any danger violence, showing great safety and generalization. Table 2 presents the performance of different methods across three generalization scenarios. 1) In the random position generalization scenario, SAGE-LLM significantly outperforms baseline methods in both task success rate and zero-danger epoch rate, with only a slightly lower success rate compared to the purely game-theoretic DDPG, while surpassing it substantially in safety. 2) In the increased obstacle quantity scenario, SAGE-LLM maintains a relatively high success rate and safety rate, whereas most safe reinforcement learning methods exhibit a sharp decline in success rate, demonstrating their vulnerability to novel generalization challenges. 3) When confronted with previously unseen cylindrical obstacles, SAGE-LLM still achieves a high safety rate, reflecting its strong zero-shot generalization capability across geometric types, which holds significant practical relevance for real-world dynamic environments. Besides,according to the error regions in Figure 3, the analysis concludes that the fluctuations in success rate and safety rate of our method are not significant.

\subsection{Experiment 3: Architectural Robustness }
\subsubsection{Experimental Setup}
This experiment is designed to evaluate the compatibility and stability of the SAGE-LLM framework when driven by various LLM backends. To provide a comprehensive assessment of architectural robustness, we selected representative models based on the following two dimensions:

Cross-Model Series: We utilized three top-tier LLMs with distinct technical lineages and optimization strategies, namely Qwen-max, DeepSeek Chat, and GLM-4.5. These models vary significantly in their pre-training corpora, alignment techniques, and architectural nuances,~\cite{sun2025speed} serving as a rigorous testbed for the generalizability of SAGE-LLM across heterogeneous logical cores.

Cross-Size Scales: Focusing on the Qwen series, we evaluated four versions spanning different parameter magnitudes: Qwen-max, Qwen3-235b, Qwen3-32b, and Qwen3-8b. This dimension aims to investigate the specific impact of model scale reduction on decision-making safety and task execution efficiency.

The scenario setup remains the same with the Randomization of Obstacle Position described in Section 5.3.

\begin{table}[h]
    \centering
    \resizebox{\linewidth}{!}{
    \begin{tabular}{cccc}
        \toprule
        Base Model      & Success Rate      & Safe Rate     & Zero-Danger-Epoch \\
        \midrule
        Qwen-max        & \textbf{80\%}     & 99.68\%                   & \textbf{85\%}      \\
        Qwen3-235b      & 70\%              & 99.11\%                   & 70\%      \\
        Qwen3-32b       & 75\%              & 99.13\%                   & 70\%      \\
        Qwen3-8b        & 50\%              & 98.55\%                   & 65\%      \\
        DeekSeek Chat   & 75\%              & 99.51\%                   & 80\%      \\
        GLM-4.5         & 75\%              & \textbf{99.71\%}          & 85\%      \\
        \bottomrule
    \end{tabular}}
    \caption{Architecture Robustness}
    \label{tab:robustness}
\end{table}

\subsubsection{Results and Analysis}

Table 2 summarizes the performance of SAGE-LLM across various backend models. The results lead to the following observations:

1) Unified Robustness Across Heterogeneous Architectures: The experimental data indicate that SAGE-LLM maintains a safe rate exceeding 99.5\% and a stable Zero-Danger-Epoch rate of approximately 80\%, regardless of whether DeepSeek Chat or GLM-4.5 is employed as the backbone. This demonstrates that the SAGE-LLM framework, by integrating the fuzzy CBF verification mechanism with the star-hierarchical graph RAG system, effectively shields the system from the inherent stochasticity and potential hallucinations of the underlying LLMs, exhibiting exceptional architectural robustness.

2) Performance Scaling and Bottlenecks Across Model Sizes: Evaluation across the Qwen series demonstrates that SAGE-LLM maintains a consistently high safety floor regardless of model scale, effectively decoupling safety from core reasoning capacity. However, task success exhibits a strong positive correlation with parameter size; while larger backbones achieve superior success rates of approximately $80\%$, the smaller model suffers significant performance degradation, with its success and Zero-Danger-Epoch rates falling to $50\%$ and $60\%$, respectively.

Analysis suggests that while the Fuzzy-CBF module successfully constrains action boundaries to prevent collisions, smaller models struggle with the complex adversarial logic and precise instruction alignment required for air combat. This indicates that while SAGE-LLM can provide provable safety for models of any size, efficient task completion still relies on the advanced reasoning and control capabilities inherent in larger-scale models.

\subsection{Experiment 4: Ablation Studies}
This section evaluates the individual contributions of the Fuzzy-CBF verification module and the Graph-RAG retrieval module to the system's overall performance.
\subsubsection{Fuzzy-CBF in Safety Assurance}
1) Safety Performance: As illustrated in Table 3, the Fuzzy-CBF module serves as a critical component for maintaining UAV flight safety. Removal of this module leads to a significant reduction in safety metrics. These results suggest that while LLMs can comprehend semantic avoidance instructions, they may lack the deterministic precision required for high-stakes dynamic environments, highlighting the necessity of physical safety constraints.

2) Performance Limitations: Comparing Row 1 and Row 2 in Table 3 reveals that although the Fuzzy-CBF only configuration achieves a superior safe rate, it suffers from a comparatively lower success rate. This indicates that a safety-only approach tends to be overly conservative, revealing a performance bottleneck in purely reactive safety mechanisms.

\subsubsection{Graph-RAG to Tactical Logic}
1) Mission Efficiency: The Graph-RAG module significantly enhances the system's tactical reasoning capabilities. By integrating structured domain knowledge, the framework enables the LLM to navigate complex pursuit scenarios more effectively, raising the Success Rate from a baseline of 50\% to approximately 80\% in the full configuration.

2) Aggressive Risks: A comparison between Row 3 and Row 4 reveals a noteworthy phenomenon: removing Graph-RAG results in a higher safe rate compared to the Graph-RAG only variant. This suggests that the Graph-RAG module may encourage aggressive maneuvers to maximize task success, demonstrating the safety risks of unconstrained retrieval-augmented reasoning.

\subsubsection{Collaborative Synergy }
The ablation study confirms that Fuzzy-CBF and Graph-RAG address different dimensions of the decision-making process, safety and efficiency, respectively. Relying on either module in isolation may lead to sub-optimal outcomes characterized by either excessive passivity or risky aggression. The SAGE-LLM architecture mitigates these individual limitations through the synergistic integration of both components. This approach achieves a balanced performance profile, maintaining a high safety standard while ensuring robust task execution.

\begin{table}[h]
    \centering
    \resizebox{\linewidth}{!}{
    \begin{tabular}{cccccc}
    
        \toprule
        Fuzzy CBF   &  Graph RAG    &  Success Rate  &  Safe Rate  & Zero-Danger-Rate    \\
        \midrule
            \checkmark  &\checkmark  & \textbf{80\%}  & \textbf{99.68\%}  & 85.00\%       \\
            \checkmark  &            & {55\%}         & 99.05\%           & \textbf{90.00\%}\\
                        &\checkmark  & {70\%}         & 97.12\%           & 45.00\%       \\
                        &            & 50\%           & 98.81\%           & 60.00\%       \\
        \bottomrule
    \end{tabular}
    }
    \caption{Ablation Test}
    \label{tab:ablation1}
\end{table}

\section{Conclusion}
This paper proposes SAGE-LLM, a train-free two-layer decision-making architecture, addressing the challenges of safety and generalization for UAV decision in dynamic and unknown environments. The framework introduces three key innovations: 1) A train-free two-layer architecture that leverages LLMs to enhance generalization capabilities for emergent threats. 2) A Fuzzy-CBF verification mechanism that provides formal physical safety boundaries for LLM-generated semantic commands. 3) A star-hierarchical graph RAG system that significantly improves tactical reasoning and precision control through structured domain knowledge retrieval. Extensive experiments demonstrate that the proposed architecture achieves a superior balance between mission efficiency and flight safety, exhibiting robust performance and cross-model compatibility in complex pursuit-evasion scenarios.

For future research, we aim to optimize the framework's computational efficiency to facilitate real-time deployment on resource-constrained embedded platforms. Furthermore, we plan to extend this architecture to accommodate multi-UAV swarm coordination and explore its broader applicability across other Embodied AI domains, such as ground robotics and autonomous driving.



\bibliographystyle{named}
\bibliography{ijcai26}

\end{document}